\documentclass[journal]{IEEEtran}
\usepackage{cite}

\ifCLASSINFOpdf
\else
\fi

\usepackage[utf8x]{inputenc} 
\usepackage{amsmath, amsthm} 
\usepackage[bottom]{footmisc}
\usepackage{cite}
\usepackage{graphicx}
\usepackage{longtable}
\usepackage{float}
\usepackage{multirow}
\usepackage{subfigure}
\usepackage{array}
\usepackage{amssymb,bm,cite,graphicx, fixmath, texdraw}
\usepackage{epsfig}
\usepackage{epstopdf}
\usepackage{subfigure} %
\usepackage{amsmath}
\usepackage{notoccite}

\usepackage{amsmath,amssymb,amsfonts}
\usepackage{algorithm}
\usepackage{algorithmicx}
\usepackage[noend]{algpseudocode}
\usepackage{graphicx}
\usepackage{textcomp}
\usepackage{xcolor}
\usepackage{etoolbox}
\usepackage[flushleft]{threeparttable}
\usepackage{ragged2e}
\usepackage{multirow}
\usepackage{listings}

\usepackage{silence}
\algrenewcommand\algorithmicrequire{\textbf{Input:}}
\algrenewcommand\algorithmicensure{\textbf{Output:}}

\makeatletter
\AfterEndEnvironment{algorithm}{\let\@algcomment\relax}
\AtEndEnvironment{algorithm}{\kern2pt\hrule\relax\vskip3pt\@algcomment}
\let\@algcomment\relax
\newcommand\algcomment[1]{\def\@algcomment{\footnotesize#1}}
\algnewcommand{\LeftComment}[1]{\Statex \(\triangleright\) #1}

\renewcommand\fs@ruled{\def\@fs@cfont{\bfseries}\let\@fs@capt\floatc@ruled
  \def\@fs@pre{\hrule height.8pt depth0pt \kern2pt}%
  \def\@fs@post{}%
  \def\@fs@mid{\kern2pt\hrule\kern2pt}%
  \let\@fs@iftopcapt\iftrue}
\makeatother

\def\BibTeX{{\rm B\kern-.05em{\sc i\kern-.025em b}\kern-.08em
    T\kern-.1667em\lower.7ex\hbox{E}\kern-.125emX}}



\usepackage{longtable}

\newcommand{\beq}{\begin{equation} \setlength\abovedisplayskip{5pt} 
\setlength\belowdisplayskip{5pt}}
\newcommand{\eeq}{\end{equation}}
\newcommand{\bea}{\begin{eqnarray}}
\newcommand{\eea}{\end{eqnarray}}

\renewcommand{\theequation}{\arabic{equation}}

\def\ifundefined{\@ifundefined}
\def\bfat{\left[ \begin{array}}
\def\emat{\end{array} \right]}
\def\bfatt{\left\{ \begin{array}}
\def\ematt{\end{array} \right.}
\def\bset{\left\{ \begin{array}}
\def\eset{\end{array} \right\}}
\def\bpar{\left( \begin{array}}
\def\epar{\end{array} \right)}

\graphicspath{{figures/}} 

\begin{document}
%
\title{EdgeConvEns: Convolutional Ensemble Learning\\for Edge Intelligence}
%
%
%

\author{Ilkay~Sikdokur, 
        \.{I}nci~M.~Bayta\c{s}, 
        and~Arda~Yurdakul
\thanks{I. Sikdokur is with the Department
of Computational Science and Engineering, Bogazici University, Istanbul, Turkey (e-mail: ilkay.sikdokur@boun.edu.tr).}
\thanks{\.{I}.~M.~Bayta\c{s} and A. Yurdakul are with the Department
of Computer Engineering, Bogazici University, Istanbul, Turkey (e-mail: inci.baytas@boun.edu.tr, yurdakul@boun.edu.tr).}}

\maketitle

\begin{abstract}
Deep edge intelligence aims to deploy deep learning models that demand computationally expensive training in the edge network with limited computational power. Moreover, many deep edge intelligence applications require handling distributed data that cannot be transferred to a central server due to privacy concerns. Decentralized learning methods, such as federated learning, offer solutions where models are learned collectively by exchanging learned weights. However, they often require complex models that edge devices may not handle and multiple rounds of network communication to achieve state-of-the-art performances. This study proposes a convolutional ensemble learning approach, coined EdgeConvEns, that facilitates training heterogeneous weak models on edge and learning to ensemble them where data on edge are heterogeneously distributed. Edge models are implemented and trained independently on Field-Programmable Gate Array (FPGA) devices with various computational capacities. Learned data representations are transferred to a central server where the ensemble model is trained with the learned features received from the edge devices to boost the overall prediction performance. Extensive experiments demonstrate that the EdgeConvEns can outperform the state-of-the-art performance with fewer communications and less data in various training scenarios.
\end{abstract}

%
\IEEEpeerreviewmaketitle

\section{Introduction}
\label{section:introduction}
Edge computing is vital for next-generation solutions in various industries such as automotive~\cite{automotive}, energy \cite{oilAndGas}, and agriculture~\cite{agriculture}. Edge intelligence offers innovative frameworks for Internet of Things (IoT) applications. Sensors are used in data collection for safe driving~\cite{automotive}, anomaly detection \cite{oilAndGas}, real-time and predictive analytics \cite{agriculture}. The data-driven problems usually require computation-intensive solutions, e.g., Deep Neural Network (DNN) training. Deep edge intelligence, where DNNs are employed on edge, has become imperative to address complex edge intelligence applications such as image classification~\cite{imageClassificationExample3}, speech recognition~\cite{speechRecognitionExample3}, and natural language processing~\cite{nlpExample3}. However, centralized training with millions of parameters is essential to attain state-of-the-art performances. In contrast, edge intelligence face problems with distributed data and limited computational resources. Consequently, two main challenges emerge in deep edge intelligence applications; data cannot be centralized due to privacy concerns and limited bandwidth, and edge devices may not supply sufficient memory and computational requirements to train complex DNNs \cite{zaidi2022unlocking}.

\begin{figure}[!t]
\centerline{\includegraphics[width=1\linewidth]{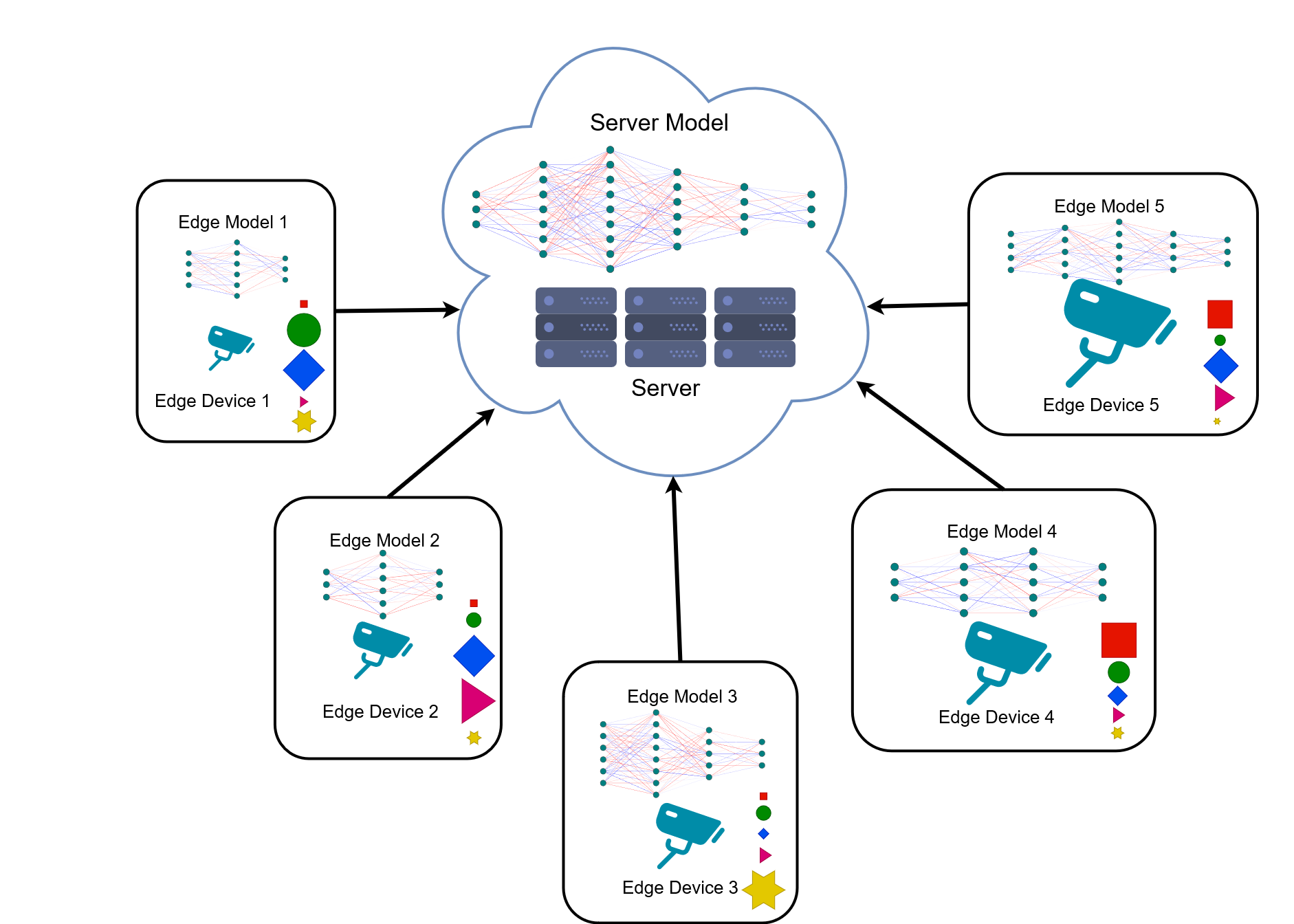}}
\caption[Deep Edge Intelligence Illustration]{Deep edge intelligence setting covered in this study. Cameras represent edge devices: processing capacity is proportional to the size. Five different colored shapes represent different classes: the bigger the size is, the more frequently it is observed by each edge device.}
\label{image:edgeIntelligenceIllustration}
\end{figure}

One prominent approach to the distributed and heterogeneous data challenge is federated learning, a popular approach in IoT solutions~\cite{fedAvg}. However, centralized and decentralized federated learning requires multiple communication rounds to update model parameters increasing the data transfer rate. Furthermore, federated learning frameworks, particularly with centralized orchestrator setup, require the clients to have similar architectures with a sufficient number of parameters to solve complex tasks. Therefore, the federated learning paradigm does not directly address the lack of computational resources at the edge of networks. Tiny Machine Learning (TinyML) is one of the recent techniques to address the low-resource device challenge. A typical TinyML framework comprises training a machine learning model on a high-performance computer, compressing the learned model to reduce memory requirements, and deploying it on an embedded system for inference \cite{zaidi2022unlocking}. Although TinyML solutions can mitigate the memory requirements, data privacy, and on-device model training challenges remain.


This study aims to improve overall prediction performance under the distributed data setup by learning to fuse the knowledge acquired from on-device training on edge with low computational resources. The proposed solution, EdgeConvEns in Figure~\ref{image:edgeIntelligenceIllustration}, combines weak learning at the edge devices with convolutional ensemble learning at the server. Unlike traditional federated learning, edge devices can train different models in the EdgeConvEns framework due to device limitations. Only the learned feature representations are transferred without the requirement for global updates. The data need not be equally available at the edge devices. Some classes can be observed by one device more frequently than others. For example, a camera viewing the right lane of a highway observes heavy vehicles more frequently than a camera on the left lane. Unlike TinyML solutions, the proposed framework trains edge models on edge devices with their local data instead of creating a compressed global model and sending it back to the edge. The contributions of this study are outlined as follows:

\begin{itemize}
\item Limited training of configurable weak models on heterogeneous edge devices is achieved on Field-Programmable Gate Array (FPGA) fabric. Autonomously generating a partially parallelized task-level streaming architecture constrained with available resources on various edge devices enables acceleration of neural network operations.
\item Variational Auto-Encoder (VAE) models are trained on the server for each edge model to generate statistically similar features to the ones learned on edge for replacing the missing features due to disconnection problems in edge intelligence. 
\item A convolutional ensemble learning scheme that fuses the underlying information in the representations learned on edge is proposed to boost overall prediction performance.

\end{itemize}
EdgeConvEns can attain satisfactory performances compared to the state-of-the-art with a much smaller number of parameters. Communication and memory utilization of the proposed framework are investigated under various data distribution and training scenarios with benchmark datasets in image classification and regression.

This work is organized as follows. Selected previous works in edge intelligence, FPGA acceleration in neural networks, federated learning and ensemble learning are analyzed in the next section. Section~\ref{section:systemStructureAndWorkflow} presents details of the proposed design and its components. Section~\ref{section:experimentalResults} contains the implementation details, experimental results and a comparative analysis with earlier works. The last section concludes this work with a review of the achievements of the study.

\section{Related Work}
\label{section:relatedWorks}

\begin{figure}[!t]
\centerline{\includegraphics[width=0.6\linewidth]{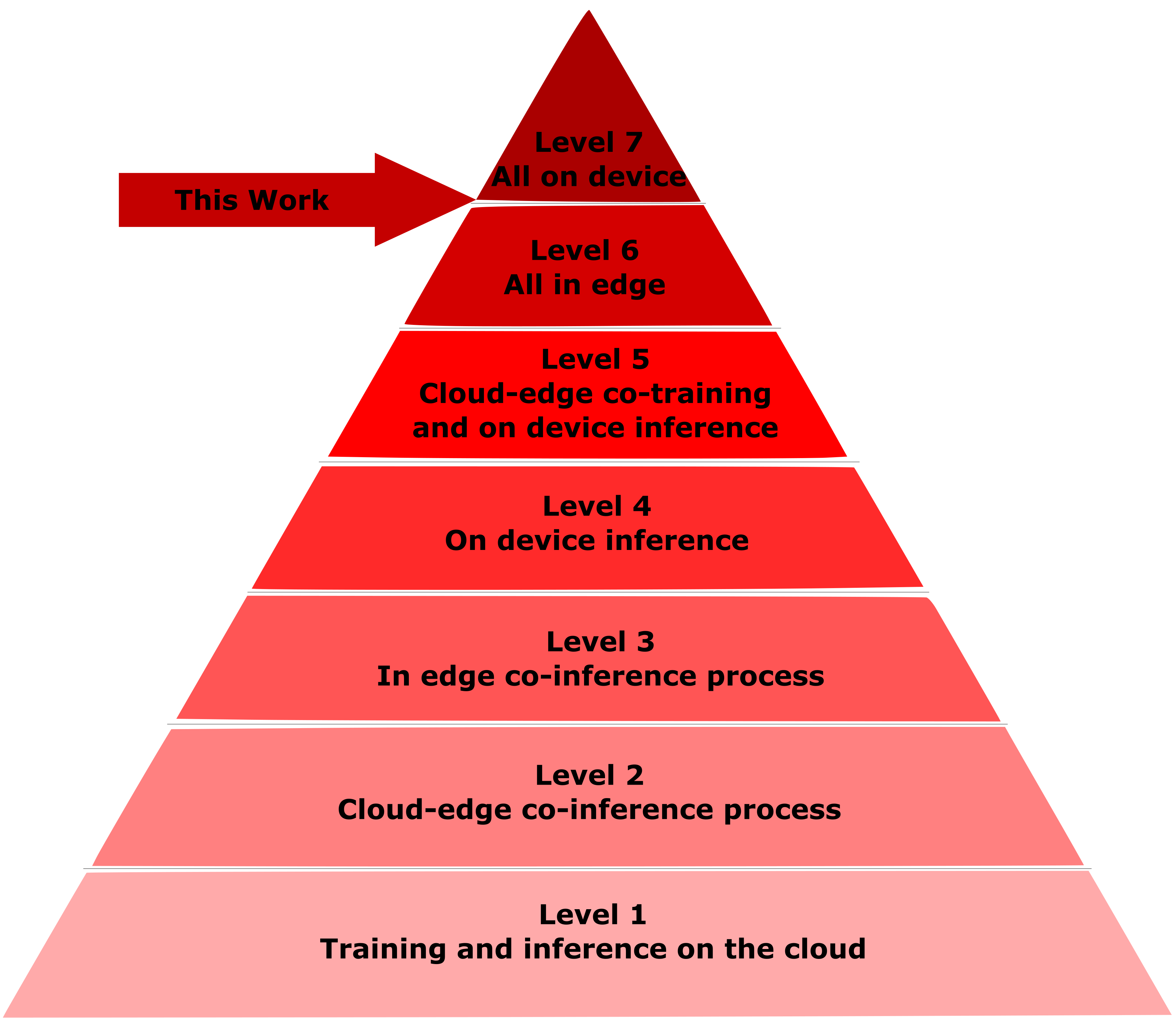}}
\caption[Edge Intelligence Levels]{Edge intelligence levels by data offloading \cite{edgeIntelligenceSurvey}.}
\label{image:EdgeIntelligencePyramid}
\end{figure}

The edge intelligence literature showcases various ways of deploying model training and inference. In some applications, training and inference are on a cloud server, whereas some applications divide training and inference into cloud server and edge, respectively. 
Zhou \textit{et al.} proposes a data offloading scale of seven levels based on the design choice of edge intelligence applications given in Figure~\ref{image:EdgeIntelligencePyramid} \cite{edgeIntelligenceSurvey}. Volume and path length of data unloading decrease as the level increases. The proposed EdgeConvEns can be placed between Levels 6 and 7 since the models on edge are independently trained, and their inference is also independently done on individual edge devices. However, the features of the inference data obtained from the edge devices are required on the server device for making the final prediction. Thus, the proposed approach results in less transmission latency of data offloading, higher data privacy, and lower communication bandwidth. EdgeConvEns can be categorized as Device-to-Edge (D2E), except that edge models do not receive any information from the server for training and inference. When the edge model training is completed on the edge device, data offload between the device and server can be less compared to other D2E approaches \cite{d2E1}.

As edge intelligence has grown more prevalent for applications like IoT, decentralized learning techniques on edge have become imperative. One of these techniques is Federated Learning (FL), a mainstream decentralized learning method commonly used in IoT. The FL methods encounter several challenges, such as the high number of communication rounds and weight transfer due to the iterative updates of local and global models~\cite{fedAvg}. In addition, all clients and the central server have the same model architecture, which dictates a minimum resource requirement for training the models on each edge device \cite{fedAvg, fedProx, fedDyn, scaffold, fedGen, fedFTG, fedDC, fedProto}. On the other hand, the EdgeConvEns is based on ensemble learning discarding the exchange of model parameters that causes a communication bottleneck. In the proposed approach, there is only a one-way transfer from edge devices to the server in which the overall prediction is boosted. In recent FL studies \cite{fedBayesianNonparametric, fedRobustFederated, fedGKT}, training is also done with either a one-way model parameter or embedding vector transfer, but EdgeConvEns yields a better prediction accuracy with a lower number of parameters according to our comparative experimental analysis.

\begin{figure*}[!ht]
\centerline{\includegraphics[width=0.65\textwidth]{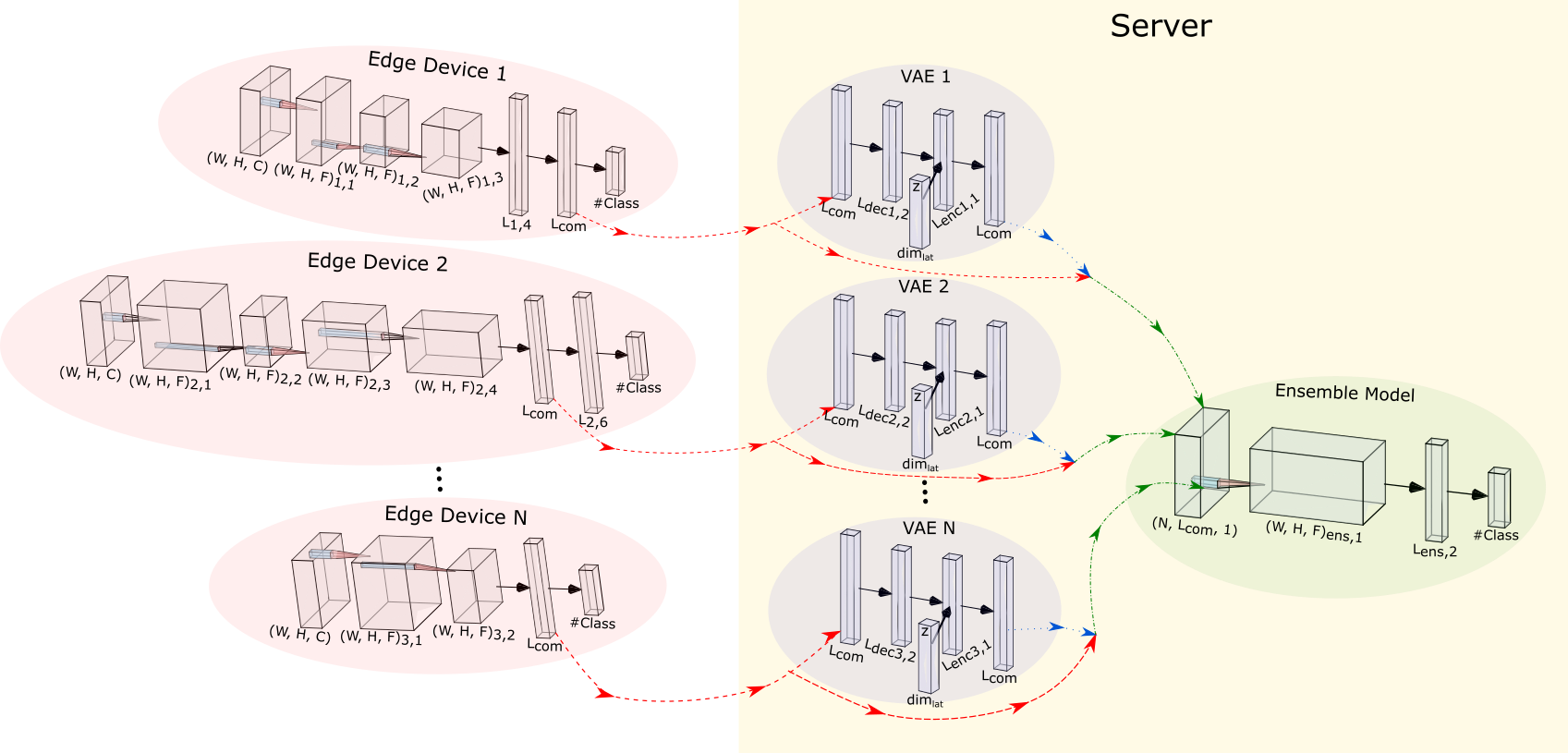}}
\caption[System Workflow]{Illustration of the system for the image classification task. Red-dashed lines denote the successful transfer of edge features. Missing features are filled by VAE outputs (blue-dashed lines). Green-dashed lines represent the complete feature set to train the CNN-based ensemble model. The system is constructed the same for regression problems.}
\label{image:System}
\end{figure*}

Ensemble methods are broadly used in deep learning. Some recent studies propose IoT frameworks based on ensemble models \cite{ensIoTExample1}, where traditional sampling strategies, such as bagging and boosting, are used. Knowledge distillation is also utilized in ensemble frameworks for edge intelligence. Such methods employ an ensemble of logit outputs and gradients of larger models to train smaller models \cite{aekd, ekd}. EdgeConvEns proposes convolution-based ensemble learning that fuses feature vectors transferred from edge for an improved classification performance compared to the traditional and knowledge distillation-based methods.

Field-Programmable Gate Arrays (FPGA) have recently been employed in edge intelligence applications to accelerate computationally expensive operations \cite{fpgaEdgeIntelligence0}. The training time of Convolutional Neural Networks (CNN) can be reduced by accelerating matrix multiplication. These studies focus on the parallelization of the calculations by certain levels only, such as fully connected layer calculations \cite{basarim}, convolution layer calculations \cite{fpgaMatrixMultAcc}, and parallelization on minibatch level \cite{fpgaBatchLevelParallel}. The proposed EdgeConvEns partially parallelizes the computations at each layer by chosen array dimensions, including minibatch
level, channel, and filter level for convolutional operations, and chosen partition level for fully connected layer operations. It also allows for the adjustment of resource utilization by changing the factor of parallelization. Some works also accelerate the training phase by binarizing \cite{fpgaAcceleration3}, quantizing and using smaller data types \cite{fpgaAcceleration5} for the weights. Even though the same optimizations can also be preferred in EdgeConvEns, the acceleration of the training is achieved without using any quantization method or smaller data types in this work.

\begin{figure}[!t]
\centering
  \subfigure[]{
    \includegraphics[width=.45\linewidth]{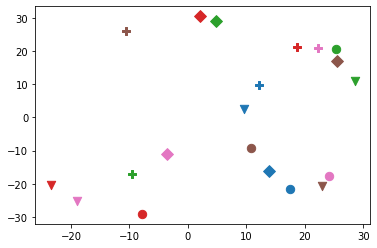}
    }
  \hspace{0.1pt}
  \subfigure[]{
    \includegraphics[width=.45\linewidth]{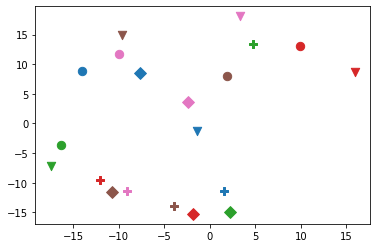}
   }
  \caption[Two-Dimensional t-SNE Distribution Plot]{Two-dimensional t-distributed stochastic neighbour embedding (t-SNE) plots of the average feature vectors extracted from 20 edge models. The feature vectors are obtained from the training set of the CIFAR-10 dataset for two different classes such as (a) automobile and (b) bird. Each symbol represents a different model.}
  \label{image:TSNETrainingDistributionCIFAR10}
\end{figure}

\section{Proposed System}
\label{section:systemStructureAndWorkflow}

EdgeConvEns, illustrated in Figure~\ref{image:System}, consists of a server and edge devices with varying hardware capacities that presumably have medium computational power. The edge models in this study are shallow neural networks with poor evaluation performances trained for image recognition and regression. The server ensemble model aims to boost the overall prediction performance by learning to fuse the edge model embeddings. The server has no computational hardware restrictions for training a complex ensemble model. The learned ensemble model makes the final predictions without broadcasting any learned model back to the edge devices.

First, we independently train the edge models with their training datasets, which are subsets acquired by random sampling with replacement from a more extensive training set of independent and identically distributed (i.i.d.) samples. Thus, edge datasets might have overlapping and distinct samples. After the training of edge models, learned embeddings of each edge's training set are transferred to the server. On the server, we first train a VAE for each edge model with respective embeddings to capture their distributions. The trained VAEs are utilized for imputing the missing embeddings by generating one from the corresponding edge's VAE. Finally, the ensemble model is trained with the training embeddings, either the original embeddings from the edge or generated from the respective edge's VAE.

\subsection{Edge Models}
\label{subsection:systemOfEdgeModels}

In the proposed framework, the edge models do not need to share the same model architecture and are trained independently. However, the dimensionality of the learned embeddings that are transferred to the server, denoted by $L_{com}$ in Figure~\ref{image:System}, should be the same since the ensemble model on the server applies convolution on the matrix of edge embeddings. The edge model denoted by $EM_i$ is trained with a subset of the training data denoted by $D_{train}^{EM_i}$, where $D_{train}$ is the main training data. After training, embeddings of $D_{train}^{EM_i}$ denoted by $F_{train}^{EM_i} \in \mathbb{R}^{L_{com}}$ is obtained. The embeddings, $F_{train}^{EM_i}$, are then transferred to the server as illustrated with red-dashed lines in Figure~\ref{image:System}.

The ensemble model expects to receive an embedding from each edge model to learn how to fuse them to improve the prediction performance. However, there may be several problems in acquiring the embedding from an edge model in real-world applications, such as connection problems between edge and server and data availability on edge. Hence, every edge model may not produce an embedding vector to transfer to the server for the same input data.

\subsection{Server Models}
\label{subsection:serverModels}
\subsubsection{Variational Auto-Encoder (VAE) Models}
\label{subsubsection:VAEModels}
This study proposes to use a generative approach to fill the necessary feature vectors not received from an edge model ${EM_i}$ to obtain the final prediction of the ensemble model. In Figure~\ref{image:TSNETrainingDistributionCIFAR10}, the dimensions of the feature vectors obtained from the layers of 20 edge models for CIFAR-10 are reduced to two by using the t-distributed Stochastic Neighbour Embedding (t-SNE) method \cite{tsne} and their average positions by edge models are plotted. It can be seen that the edge outputs to be used in the ensemble model follow different distributions. EdgeConvEns leverages VAE models for imputing the missing embeddings. VAE models are trained and deployed on the server for each edge model to generate embeddings sampled from a similar distribution as the edge embeddings'. In Figure~\ref{image:System}, the red-dashed line from our edge device to the VAE model shows the transfer of the edge embeddings $F_{train}^{EM_i}$. Each VAE model, $VAE_i$, is independently trained with respective edge embeddings, $F_{train}^{EM_i}$, to capture the edge models' embedding distributions. For this purpose, embeddings received from the edge models are encoded into a distribution over latent space. When a random sample from the encoder's distribution is decoded to reduce the reconstruction error, the VAE's decoder starts exhibiting a generative nature. Thus, the trained VAE's decoder can be used to generate embeddings, $VAE_i^{dec}(z)$, from a random variable in the latent space, $z \in \mathbb{R}^{d}$.

\begin{figure}[!t]
\centerline{\includegraphics[width=.7\linewidth]{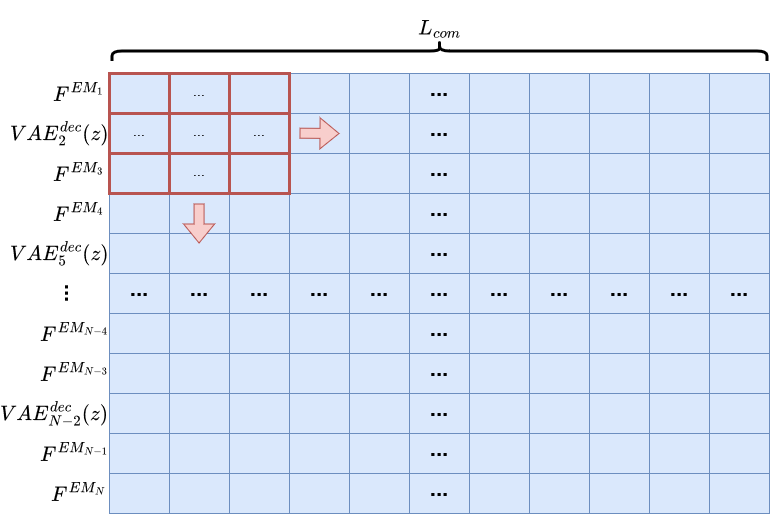}}
\caption[Convolutional Ensemble Operation]{Illustration of the convolutional ensemble operation for a single image. $F^{EM_i}$: the feature vectors transferred from the $i$-th edge model. $VAE^{dec}_i(z)$: the feature vectors generated by $i$-th VAE model for the same image.}
\label{image:convolutionalEnsemble}
\end{figure}

\subsubsection{Ensemble Model}
\label{subsubsection:ensembleModel}

Since the weak models on edge devices cannot yield state-of-the-art performances, a convolutional ensemble approach shown by the green model in Figure~\ref{image:System} is proposed for learning a fusion of embeddings to improve the final predictive performance. Let $K_{train}$ contain the indices of the samples for the full training dataset $D_{train}$ and $K_{train}^{EM_i}$ contain the indices for the edge model's ($EM_i$) training dataset, $D_{train}^{EM_i}$. Since $D_{train}^{EM_i} \subseteq D_{train}$, it can be deduced that $K_{train}^{EM_i} \subseteq K_{train}$. The training dataset for the ensemble model, $D^{Ens}_{train}$, is constructed as follows.   
\beq \label{equation:fillingFunction}
{D^{Ens}_{train}[k, i]  = \left\{\begin{matrix}
F_{train}^{EM_i}  & if \ k \in K_{train}^{EM_i} \\ 
VAE_i^{dec}(z) & if \ k \notin K_{train}^{EM_i}
\end{matrix}\right.
}.
\eeq Thus, the ensemble model's training data, $D^{Ens}_{train}$, of size $n(D_{train}) \times N \times L_{com}$ is constructed by stacking the embeddings. Figure~\ref{image:System} shows the ensemble training data construction with blue and green-dashed lines.

Figure~\ref{image:convolutionalEnsemble} illustrates the matrix formed by stacking the representations obtained from either edge models or VAE for a single sample. The red two-dimensional frame shows the convolutional kernel to extract a unified representation of each sample. The motivation behind the convolutional ensemble is to learn a fusion that complements each edge model's information by investigating the inter-edge interactions in the embeddings. Hence, the kernel size directly affects the prediction performance. In this study, the architecture of the ensemble model comprises a convolutional layer with a kernel of size $(\frac{N}{2}, \frac{L_{com}}{2})$ and a hidden fully-connected layer.

\section{Experimental Results}
\label{section:experimentalResults}
\subsection{Implementation}
\label{subsection:implementationDetails}

\subsubsection{Partially Parallelized Streaming in FPGA-based Edge Devices}
\label{subsubsection:implementationTaskLevelStreaming}

Inference and training of DNNs heavily require matrix multiplications. Acceleration of matrix multiplication is performed by the proposed task-level parallelism, which allows the designer to choose the level of acceleration to ensure targeted hardware utilization. The proposed method transfers data using \textit{data stream} method~\cite{datastream} where the data flow from the source to the destination module via First-In-First-Out (FIFO) buffers. FIFO buffers allow the consumer module to use the data when the source module sends the data providing accelerated computations. The data interface is chosen as AXI4-Stream~\cite{axis} for all modules. The task-level streaming method has constraints, such as reading and writing an array element only once, in addition to the in-order access to the buffers.

In EdgeConvEns, all of the arrays used in training are manually tiled in preprocessing phase by the factors decided by the designer. Then, the arrays are partitioned into tiles during synthesis by the factorized part of the chosen dimension. The factors need to be decided to fit the edge model into the chosen edge device by the designer. Assume that the matrices $M_1[X][Y]$ and $M_2[Y][Z]$ are aimed to be multiplied into the result matrix $M_3[X][Z]$. It can be seen that the $Y$ is the common dimension of multiplication. Assume that $Y = f*y$ where $f$ and $y$ are positive integers. Then, the matrix multiplication can be written as \beq
M_{3_{i, j}} = \sum_{k=1}^{f} \sum_{l=1}^{\frac{Y}{f}}  M_1[i][k][l] * M_2[k][l][j].\eeq
In this form, the matrices $M_1$ and $M_2$ are manually tiled into the $f$ factor on the $Y$ dimension. After this modification, the follow-up arrays can be automatically partitioned by the new factor dimension $f$. Then, the partitioned independent memory parts are streamed through the system in FIFO buffers.

\begin{figure}[!t]
\centerline{\includegraphics[width=0.95\linewidth]{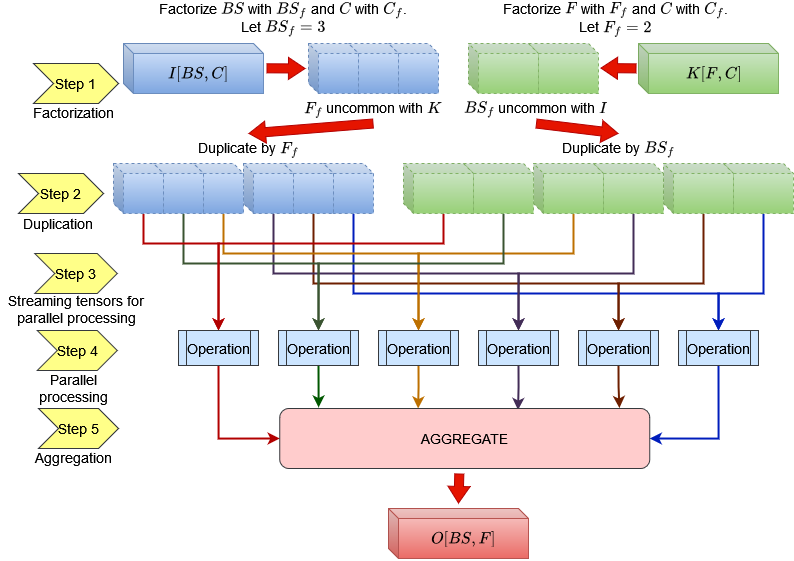}}
\caption[Array Duplication and Tiling]{Illustration of duplication, operation, and aggregation of arrays in dataflow parallelization.}
\label{image:DuplicateOperateAggregate}
\end{figure}

In Figure~\ref{image:DuplicateOperateAggregate}, the factorization operation used in EdgeConvEns is illustrated. In the first step, the tensors $I$ and $K$, are factorized by manually selected factors, namely $BS_f$ and $F_f$. Calculation operations require multiple uses of the same data which violates data streaming; therefore, required parts are duplicated in the second step. In the third step, the duplicated tensors are streamed for parallel computation. In the fourth step, the aimed calculations are done in parallel. Finally, the acquired partial results are aggregated into the result tensor in the last step.



In this study, training of edge models is implemented on Xilinx FPGAs using high-level synthesis (HLS) \cite{hls}. Vitis HLS offers several directives which enable array partitioning, streaming, and parallel computation. In the algorithm, completely partitioning the manual tiles is executed with the "$\#$pragma HLS ARRAY$\_$PARTITION complete" command where the "variable=" parameter specifies the array and the "dim=" parameter is used for the dimensions to partition. The arrays are also streamed inside the module, and pipelining is applied with the "$\#$pragma HLS DATAFLOW" command. In addition, the task-level parallelism on computations is applied with the "$\#$pragma HLS UNROLL" command. The same method applies to fully-connected layer computations with appropriate inputs and outputs.

The mini-batch size and the input channel size, convolution filters, and length of the fully connected layers need to be manually factorized into tiles for CNN calculations. The other factorizable dimension of the array is automatically factorized based on the manual factorization of the weight of the previous layer. The proposed Algorithm~\ref{algorithm:automation} automatically sets the tiling factor of the layer on the corresponding dimension based on the manually tiled dimension of the previous layer. In the algorithm, dimensions of the $i$-th convolution kernel are denoted by $[F_f^i, C_f^i][F_p^i, C_p^i][KW^i][KH^i]$ for $F^i=F_f^i \times F_p^i$ number of filters and for $C^i=C_f^i \times C_p^i$ number of channels. Dimensions for $i$-th fully connected layer weights are denoted by $[L1_f^i, L2_f^i][L1_p^i, L2_p^i]$ as $L1^i=L1_f^i \times L1_p^i$ and $L2^i=L2_f^i \times L2_p^i$. The $i$-th convolutional layer output dimensions are $[BS_f, C_f^i][BS_p, C_p^i][W^i][H^i]$, where $BS = BS_f \times BS_p$ is the minibatch size. The $i$-th fully connected layer outputs dimensions are $[BS_f, L2_f^i][BS_p, L2_p^i]$.

\begin{algorithm}[!t]
\begin{algorithmic}
    \Require Batch of images: $I[BS_f, C_f][BS_p, C_p][W][H]$, \\
    Edge model: $EM$ \\
    Manual factorizations of layers: $f$

    \For {$i<=n(l)$} \Comment{Forward Pass}
        \If {$l^{i} = $ Convolution}
            \If {$i=1$}
                \State {$dim(w^{i}) = [f^{i}, C_f][\frac{F^i}{f^{i}}, \frac{C^i}{C_f}][KW^i][KH^i]$}
            \ElsIf {$i>1$}
                \State {$dim(w^{i}) = [f^{i}, f^{i-1}][\frac{F^i}{f^{i}}, \frac{F^{i-1}}{f^{i-1}}][KW^i][KH^i]$}
            \EndIf
            \State {$dim(o^{i}) = [BS_f, f^{i}][BS_p, \frac{F^i}{f^{i}}][W^i][H^i]$}

        \ElsIf {$l^{i} = $ Fully Connected Layer}
            \If {$i=1$}
                \State {$dim(w^{i}) = [C_f, f^{i}][\frac{L1^i}{C_f}, \frac{L2^i}{f^{i}}]$}
            \ElsIf {$i>1$}
                \State {$dim(w^{i}) = [f^{i-1}, f^{i}][\frac{L1^i}{f^{i-1}}, \frac{L2^i}{f^{i}}]$}
            \EndIf
            \State {$dim(o^{i}) = [BS_f, f^{i}][BS_p, \frac{L2^i}{f^{i}}]$}
        \EndIf
    \EndFor
    \\
    \For {$i<=n(l)$} \Comment{Backward Pass}

        \If {$l^{i} = $ Fully Connected Layer}
            \State {$dim(\partial \mathcal{L}^{i}) = [BS_f, f^{i}][BS_p, \frac{L2^i}{f^{i}}]$}
            \State {$dim(gr^{i}) = [f^{i-1}, f^{i}][\frac{L1^i}{f^{i-1}}, \frac{L2^i}{f^{i}}]$}

        \ElsIf {$l^{i} = $ Convolution}
            \State {$dim(\partial \mathcal{L}^{i}) = [BS_f, f^{i}][BS_p, \frac{L2^i}{f^{i}}]$}
            \State {$dim(gr^{i}) = [f^{i}, f^{i-1}][\frac{F^i}{f^{i}}, \frac{F^{i-1}}{f^{i-1}}][KW^i][KH^i]$}
        \EndIf
    \EndFor
\end{algorithmic}
\caption[Automation Process of the Tiling]{Automation process of the tiling.}
\label{algorithm:automation}
\end{algorithm}

In Algorithm~\ref{algorithm:automation}, $l$ denotes the layers, $w$ and $o$ represent the weights and the outputs of the corresponding layers. $\partial \mathcal{L}$ stands for the loss sent back from the next layer, and $gr$ shows the corresponding gradients of the layers. Here, $dim()$ outputs the dimensions of the corresponding array, and $f$ is the manually selected tiling factorizations.

\subsubsection{Experimental Setup}
\label{subsubsection:implementationExperimentalSetup}

As mentioned in Section~\ref{section:introduction}, the data is assumed to be distributed as some edge devices observe some classes more frequently than others. Two metrics, namely $\alpha$ and $\delta$, are defined to represent this behavior. The $\alpha$ parameter denotes the minimum percentage of the whole training data of classes used in the training edge models. For example, when $\alpha=0.25$, then at least 25\% of the training data of each class is randomly sampled for each edge device. The second metric $\delta$ denotes the maximum discrepancy rate between the percentage of sampled training data and sampled test data for a class. For example, if 10\% of the whole training data of class 1 is used for training an edge model and $\delta = 0.5$, then the percentage of test data of class 1 that the edge model accesses is between 5\% and 15\%. The experiments are conducted by taking random subsets of the whole train and test data. $K^{EM_i}_{train}$ is chosen as $\bigcup_{i=1}^c \{ k_i \subset K_{train},  \frac{\left | k_i \right |}{\left |K_{train_i}\right |} = X\sim \mathcal{U}_{[\alpha, 1]} $ and $k_i$ is chosen randomly$\}$. In addition, $K^{EM_i}_{test}$ is chosen as $\bigcup_{i=1}^c \{ k_i \subset K_{test},  \frac{\left | k_i \right |}{\left |K_{test_i}\right |} = max\{min\{\left | K_{train_i} \right | * (1+X), 1 \}, 0\}$ and $ X\sim \mathcal{U}_{[-\delta, \delta]} \}$. Different values of $\alpha$ and $\delta$ remarkably change the class data percentage for train and test data of edge devices, which affects the overall prediction accuracy with timing and memory requirements.

The classification experiments are conducted on four different image classification benchmark datasets, CIFAR-10 \cite{cifar10}, CIFAR-100 \cite{cifar10}, MNIST \cite{mnist}, and Fashion MNIST \cite{fasmnist}. The regression performance of the method is evaluated on Boston Housing \cite{bostonHousing}, California Housing \cite{californiaHousing}, and Pecan Street \cite{pecanStreet} datasets.

Experiments on classification utilize train and test data randomly generated for each edge device based on $\alpha$ and $\delta$ values. The continuous values of the targets in the regression datasets are divided into ten groups in training data, where all groups have the same density of data to mimic the classification behavior. Test data are also grouped based on the training grouping. After creating classes for regression datasets, $\alpha$ is set to 0.05 in the regression experiments.

\subsubsection{Edge and Server Models}
\label{subsubsection:implementationEdgeModels}

In the experiments, each weak edge model is arbitrarily designed. For the image classification task, the networks are built with structures such as one convolutional layer with $5\times5$ kernel size and $f_e$ filters, one $2\times2$ max pooling layer, one convolutional layer with $5\times5$ kernel size and $f_e$ filters, one or two Fully Connected Layers (FCL) with 64-dimensional output, and an output FCL with c-dimensional output where $f_e\in\left \{1,2,4  \right \}$ and $c$ is the number of classes. For the regression task, the networks are built with structures such as one or two Fully Connected Layers (FCL) with 64-dimensional output and an output FCL with 1-dimensional output.

Also, the training data and training epochs are randomly sampled for each edge model. The minimum, mean and maximum statistics of the training epoch, model accuracy, and the total number of model parameters are given in Table~\ref{table:edgeModelSpecifications} for each classification dataset. It can be seen that the accuracy values of the edge models vary quite a lot because edge models differ in structure, and some edge models are trained for a few epochs. Due to their relatively small and shallow structures, the edge models do not produce a good accuracy. 

The edge models are trained independently from each other and the models that are on the server. The cross-entropy, given as $-\sum_{i=1}^{c} y_i log(\hat{y}_i)$, and mean squared error (MSE), given as $\frac{1}{n}\sum_{i=1}^{n} (y_i - \hat{y}_i)^2$, are used as the loss functions during edge model training for image classification and regression problems, respectively. In the loss functions, $y_i$ and $\hat{y}_i$ denote the ground truth label and the prediction for the $i$-th class out of $c$ classes for cross-entropy loss, where they denote $i$-th prediction and observation in the $n$-sized sample for MSE loss, respectively. The loss function used in the training of VAE models is given as $\left \| F_{train}^{EM_i}-VAE_i^{dec}(z) \right \|^2 + KL[\mathcal{N}(\mu _{enc}, \sigma _{enc}), \mathcal{N}(0, 1)]$, where $z$ denotes a random sample vector from the latent space with the distribution $\mathcal{N}(\mu _{enc}, \sigma _{enc})$, $\mu _{enc}$ and $\sigma _{enc}$ denote mean and standard deviation terms of the encoding, $KL$ denotes the Kullback-Leibler (KL) divergence and $\mathcal{N}$ denotes normal distribution. The ensemble model is trained with the same loss functions as the edge models. It should be highlighted that the edge and ensemble models are implemented using float (32-bit) datatype. ADAM optimizer is used to train VAE and ensemble models with a learning rate of $10^{-4}$. For edge models, Stochastic Gradient Descent (SGD) optimizer  is used with $10^{-4}$ learning rate.

\begin{table}[!t]
\centering
\caption{Edge model properties.}
\label{table:edgeModelSpecifications}
\resizebox{\columnwidth}{!}{
\begin{tabular}{|c|ccc|ccc|ccc|}
\hline
\multirow{2}{*}{\textbf{Dataset}}                       & \multicolumn{3}{c|}{\textbf{\begin{tabular}[c]{@{}c@{}}Number of\\ Training Epochs\end{tabular}}} & \multicolumn{3}{c|}{\textbf{\begin{tabular}[c]{@{}c@{}}Number of\\ Parameters\\ (thousand)\end{tabular}}} & \multicolumn{3}{c|}{\textbf{\begin{tabular}[c]{@{}c@{}}Accuracy\\ (\%)\end{tabular}}} \\ \cline{2-10} 
                                                        & \multicolumn{1}{c|}{\textbf{Min}}  & \multicolumn{1}{c|}{\textbf{Mean}} & \textbf{Max} & \multicolumn{1}{c|}{\textbf{Min}}        & \multicolumn{1}{c|}{\textbf{Mean}}        & \textbf{Max}       & \multicolumn{1}{c|}{\textbf{Min}} & \multicolumn{1}{c|}{\textbf{Mean}} & \textbf{Max} \\ \hline
CIFAR-10                                                & \multicolumn{1}{c|}{11}            & \multicolumn{1}{c|}{32}            & 49           & \multicolumn{1}{c|}{7}                  & \multicolumn{1}{c|}{25}                  & 44                & \multicolumn{1}{c|}{10.36}        & \multicolumn{1}{c|}{37.46}         & 51.37        \\ \hline
CIFAR-100                                               & \multicolumn{1}{c|}{10}            & \multicolumn{1}{c|}{25}            & 46           & \multicolumn{1}{c|}{13}                  & \multicolumn{1}{c|}{25}                  & 50                & \multicolumn{1}{c|}{4.53}         & \multicolumn{1}{c|}{8.43}         & 17.23        \\ \hline
MNIST                                                   & \multicolumn{1}{c|}{12}            & \multicolumn{1}{c|}{24}            & 45           & \multicolumn{1}{c|}{4}                  & \multicolumn{1}{c|}{15}                  & 21                & \multicolumn{1}{c|}{93.57}        & \multicolumn{1}{c|}{96.74}         & 97.94        \\ \hline
\begin{tabular}[c]{@{}c@{}}Fashion\\ MNIST\end{tabular} & \multicolumn{1}{c|}{10}            & \multicolumn{1}{c|}{32}            & 47           & \multicolumn{1}{c|}{4}                  & \multicolumn{1}{c|}{18}                   & 21                & \multicolumn{1}{c|}{71.76}        & \multicolumn{1}{c|}{79.91}         & 90.16        \\ \hline
\end{tabular}
}
\end{table}

\subsection{Training Scenarios}
\label{subsection:TrainingScenarios}

The training of the proposed system can be done in three different scenarios based on the feature vector transfer scheme from the edge to the server. These three scenarios bring advantages and disadvantages to the training process and stand on a trade-off between training time, memory requirement, and overall accuracy. The appropriate scenario can be chosen according to the device choice and specifications of the transfer medium. In all scenarios, $Ep^{EM}$, $Ep^{VAE}$, and $Ep^{Ens}$ denote the training epochs of edge models, VAE models, and the ensemble model, respectively.

\subsubsection{Abundant Memory on the Server}
\label{subsubsection:trainingScenario1}
Scenario 1 is considered when all feature vectors obtained from the inference of training data are transferred from the edge models to the server at once. Therefore, VAE training and ensemble learning can start once the feature vector transfer is completed. In this scenario, the memory requirements on the server side are the highest among the three scenarios since all the feature vectors must be stored on the server device. It also achieves the highest accuracy among all the scenarios discussed in this study. 

Firstly, edge models are trained independently using their training data for $Ep^{EM}$ epochs. After their training, feature vectors, $F_{train}^{EM_i}$, with $L_{com}$ length are obtained by inference. After the inference, the feature vectors are transferred to the server in only one transfer. When the transfer is done, the VAE model of each edge model is trained for $Ep^{VAE}$ epochs. Then, missing feature vectors are replaced via these trained VAE models for each edge model. Ultimately, the final ensemble model is trained using all feature vectors received and replaced.

\subsubsection{Limited Memory on the Server}
\label{subsubsection:trainingScenario2}
In Scenario 2, only a mini-batch of ensemble training data consisting of feature vectors is sent to the server. This scenario remarkably reduces the memory requirement on the server side. The total ensemble training epoch $Ep^{Ens}$ is divided by a factor $Ep^{Ens}_d$. The stored mini-batch is repeatedly used in training for $\frac{Ep^{Ens}}{Ep^{Ens}_d}$ epoch. Every mini-batch is transferred to the server for $Ep^{Ens}_d$ times. This scenario requires less data load per transfer but more communication.

Training the ensemble learner using the same mini-batch repeatedly causes bias in training hence, degrading the accuracy. However, this degradation can be alleviated to a certain level by decreasing the repeated use of a mini-batch in training. For example, increasing the number of $Ep^{Ens}_{d}$ alleviates such bias in training at the cost of increasing the number of transfers. The training of VAE models is also deteriorated by low $Ep^{Ens}_{d}$. It yields lower accuracy in the feature generation than in Scenario 1. 

A special case of Scenario 2, where $Ep^{Ens} = Ep^{Ens}_d$, is named Scenario 3. In this case, a mini-batch is used in training for one epoch at one communication. On the server, extra storage memory is not required for keeping the mini-batch for consecutive use in training. Since each mini-batch is used in training for one epoch, the accuracy acquired is close to the first scenario. The disadvantage of this scenario is the increased number of one-way communication.

\subsection{Quantitative Analysis}
\label{subsection:quantitativeAnalysis}

For the experiments, $L_{com}$ is set to 64 and edge models are trained for 30 epochs. The ensemble model is trained for 100 epochs. The transfer rate between the edge and server is assumed to be 450 Mbps. Memory and communication evaluations are done for the CIFAR-10 dataset. Moreover, 60\%, 70\%, and 87\% of the full training set are used when $\alpha$ values are set to 0.05, 0.3, and 0.7 for memory and communication evaluations. The VAE and the ensemble model architecture are the same for all datasets and training scenarios. VAE model consists of an FCL with 64 neurons in encoder and decoder parts and the latent space is taken as $\mathbb{R}^{32}$. Total number of parameters used for the encoder and decoder parts of each $VAE_i$ is 260K. VAE models are trained for 50 epochs. Ensemble model consists of 64 $\frac{N}{2} \times \frac{L_{com}}{2}$ kernels and an FCL with 64 neurons. The ensemble model has 182K parameters for all datasets and training scenarios. The accuracy values are given for $\alpha=0.05$ and $\delta=0$ unless mentioned otherwise.

The FPGA experiments are conducted on Xilinx Artix-7 AC701 Evaluation Platform using CIFAR-10 dataset. The synthesis and implementation of the system are done using Vitis HLS 2022.2 and Vivado 2022.2 platforms. The VAE and ensemble model experiments are conducted on GeForce RTX 2080 Ti device. 

Firstly, the effectiveness of the proposed partial task-level streaming implemented on FPGA devices is presented. Different model layers are parallelized in the experiments with the given parallelization factors. The training is done with mini-batches of size two. The results can be seen in Table~\ref{table:dataflowUtilization}. The structure of the edge model used in this experiment is one convolutional layer with $5\times5$ kernel size and two filters, one $2\times2$ max pooling layer, one convolutional layer with $5\times5$ kernel size and two filters, one Fully Connected Layers (FCL) with 64-dimensional output, and an output FCL with 10-dimensional output. In the table, "Without Optimizations" is the training of the model without any task-level parallelism applied. $BS_f$ and $F_f$ denote the parallelization factors of the mini-batch dimension and the filter dimension of the data for edge training, respectively. Therefore, $BS_{f=x}$-$F_{f=y,z}$ points to that mini-batch are parallelized by factor $x$, the first convolutional operation is parallelized by factor $y$, and the second convolutional operation is parallelized by factor $z$. The utilization values are taken from the post-implementation phase. Latency values are taken from the C/RTL Cosimulation results. Experimental results show that task-level streaming yields remarkable acceleration even without factorizing different layer calculations. Also, parallelizing different layers with different factors shows changing effects on acceleration and resource utilization. These results show that the proposed method can be configured for devices with various hardware capacities.

\begin{table}[!t]
\centering
\caption{Latency and resource utilization of edge models.}
\label{table:dataflowUtilization}
\resizebox{\columnwidth}{!}{
\begin{tabular}{|c|c|c|c|c|c|c|c|}
\hline
\textbf{Configuration} &
  \textbf{\begin{tabular}[c]{@{}c@{}}Latency \\ (ms)\end{tabular}} &
  \textbf{\begin{tabular}[c]{@{}c@{}}BRAM\\ (\%)\end{tabular}} &
  \textbf{\begin{tabular}[c]{@{}c@{}}DSP\\ (\%)\end{tabular}} &
  \textbf{\begin{tabular}[c]{@{}c@{}}FF\\ (\%)\end{tabular}} &
  \textbf{\begin{tabular}[c]{@{}c@{}}LUT\\ (\%)\end{tabular}} &
  \textbf{\begin{tabular}[c]{@{}c@{}}LUTRAM\\ (\%)\end{tabular}} & 
  \textbf{\begin{tabular}[c]{@{}c@{}}IO\\ (\%)\end{tabular}}\\ \hline
\begin{tabular}[c]{@{}c@{}}Without\\ Optimizations\end{tabular}     & 194 & 20 & 6 & 3 & 6 & 1 & 39 \\ \hline
\begin{tabular}[c]{@{}c@{}}EdgeConvEns\\ $BS_{f=1}$-$F_{f=1,1}$\end{tabular} & 90  & 64 & 13 & 6 & 11 & 2 & 61 \\ \hline
\begin{tabular}[c]{@{}c@{}}EdgeConvEns\\ $BS_{f=1}$-$F_{f=1,2}$\end{tabular} & 72  & 69 & 17 & 7 & 14 & 3 & 61 \\ \hline
\begin{tabular}[c]{@{}c@{}}EdgeConvEns\\ $BS_{f=1}$-$F_{f=2,1}$\end{tabular} & 51  & 75 & 17 & 7 & 14 & 3 & 61 \\ \hline
\textbf{\begin{tabular}[c]{@{}c@{}}EdgeConvEns\\ $BS_{f=2}$-$F_{f=1,1}$\end{tabular}} & \textbf{46}  & \textbf{76} & \textbf{22} & \textbf{9} & \textbf{18} & \textbf{4} & \textbf{86} \\ \hline
\end{tabular}
}
\end{table}


The importance of replacing the missing feature vectors with appropriate values can be seen in Table~\ref{table:missingAndZeroFillingAccuracy}. The dataset used in this experiment is CIFAR-10. The average edge model accuracy is 40\%, and 20 edge models are used. The results show that replacing the missing feature vectors with trivial values such as zero, the mean, and the maximum element of the successfully transferred feature vectors detriments the overall accuracy of the ensemble model. The feature vectors generated with VAE produce remarkably more accurate results in terms of overall ensemble accuracy.

\begin{table}[!t]
\centering
\caption{Accuracy with different filling methods.}
\label{table:missingAndZeroFillingAccuracy}
\begin{tabular}{|c|ccc|}
\hline
\multirow{2}{*}{\textbf{\begin{tabular}[c]{@{}c@{}}Replacement\\ Method\end{tabular}}} & \multicolumn{3}{c|}{\textbf{Accuracy (\%)}} \\ \cline{2-4} 
 & \multicolumn{1}{c|}{\textbf{\begin{tabular}[c]{@{}c@{}}$\boldsymbol{\alpha = 0.3}$\\ $\boldsymbol{\delta = 0.2}$\end{tabular}}} & \multicolumn{1}{c|}{\textbf{\begin{tabular}[c]{@{}c@{}}$\boldsymbol{\alpha = 0.5}$\\ $\boldsymbol{\delta = 0.5}$\end{tabular}}} & \textbf{\begin{tabular}[c]{@{}c@{}}$\boldsymbol{\alpha = 0.7}$\\ $\boldsymbol{\delta = 0}$\end{tabular}} \\ \hline
With Zero & \multicolumn{1}{c|}{62} & \multicolumn{1}{c|}{58} & 58 \\ \hline
With Mean & \multicolumn{1}{c|}{52} & \multicolumn{1}{c|}{54} & 57 \\ \hline
With Max & \multicolumn{1}{c|}{35} & \multicolumn{1}{c|}{42} & 52 \\ \hline
\textbf{With VAE} & \multicolumn{1}{c|}{\textbf{82}} & \multicolumn{1}{c|}{\textbf{73}} & \textbf{63} \\ \hline
\end{tabular}
\end{table}

\begin{figure}[!t]
\centering
  \subfigure[]{
    \includegraphics[width=0.95\linewidth]{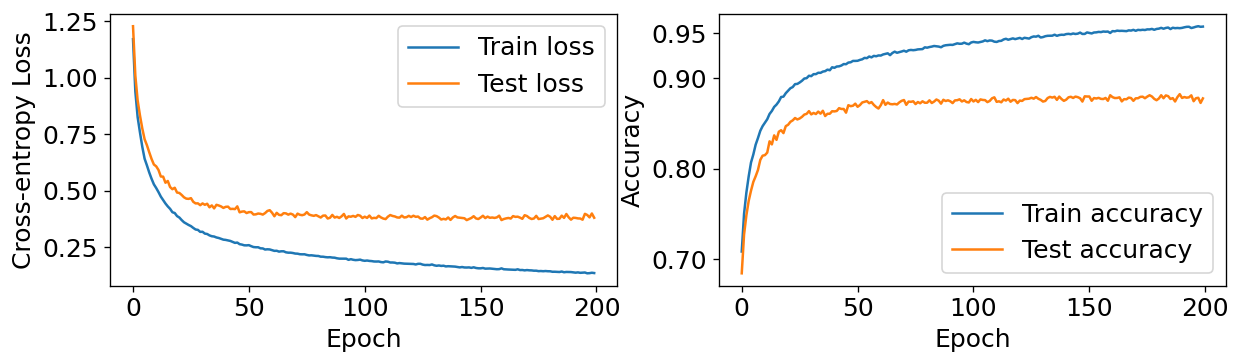}
    }
   \\
  \subfigure[]{
    \includegraphics[width=0.95\linewidth]{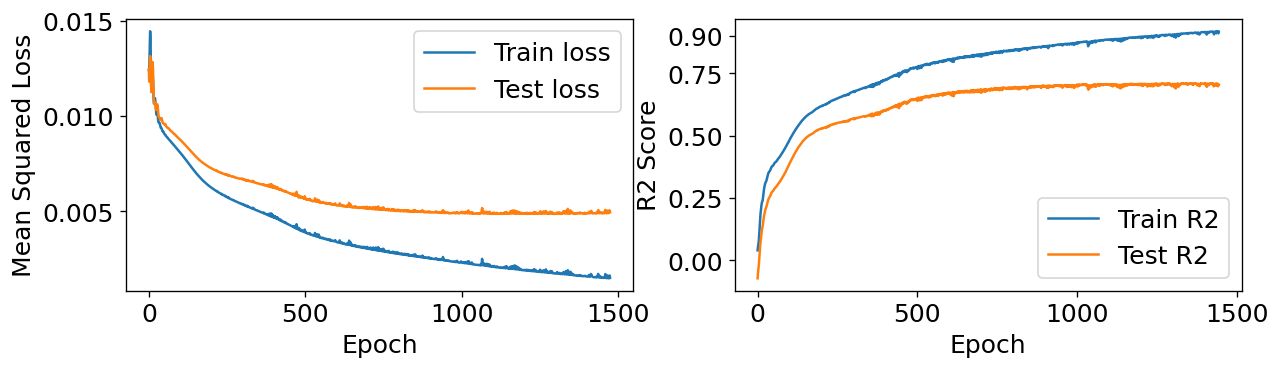}
   }
  \caption[Ensemble Model Loss and Accuracy Plot]{The loss plots of the ensemble model trained for (a) CIFAR-10 dataset for image classification, and (b) Pecan Street dataset for regression tasks.}
  \label{image:ensembleLoss}
\end{figure}

Examining the loss and accuracy plots of the training of the proposed ensemble model is essential. Figure~\ref{image:ensembleLoss} shows that the ensemble model converges to a solution as the training continues. The plots are given for $\alpha = 0.05$ and $\delta = 0.5$ for CIFAR-10, and Pecan Street datasets on Scenario 1.

\begin{figure}[t!]
\centering
  \subfigure[]{
    \includegraphics[width=0.45\linewidth]{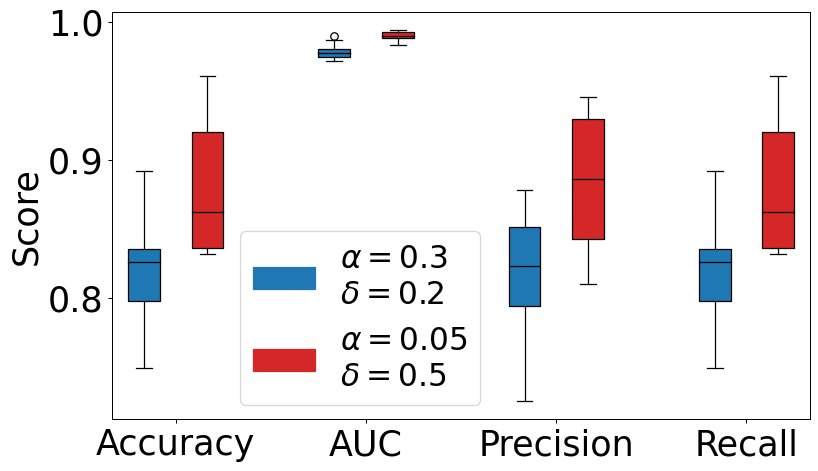}
    }
  \hspace{0.1pt}
  \subfigure[]{
    \includegraphics[width=0.45\linewidth]{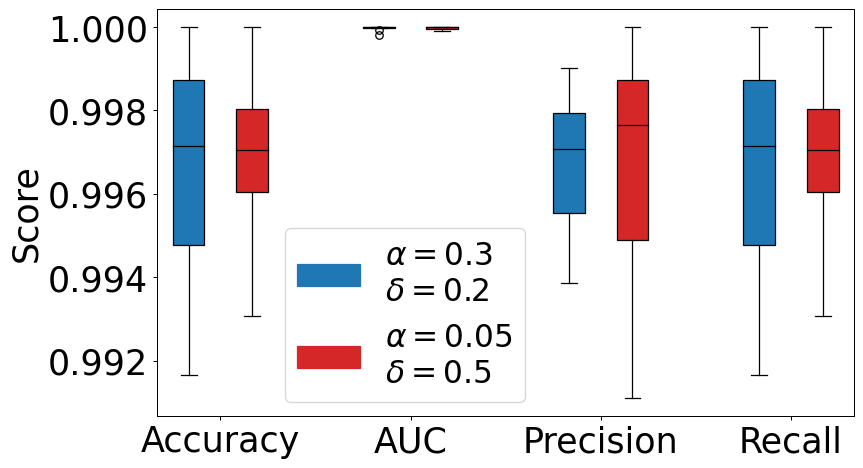}
   }
  \caption[Ensemble Model Performance Metrics]{Performance metrics of the ensemble model on different classes where $N=20$. (a) CIFAR-10, and (b) MNIST.}
  \label{image:performanceMetrics}
\end{figure}

Figure~\ref{image:performanceMetrics} shows the box plot of the performance metrics of accuracy, the area under the ROC curve, precision, and recall values calculated for each class in test data for CIFAR-10 and MNIST datasets. The values are separately calculated based on one-hot encodings for the classes. Even though the edge models are not fully trained for all classes, the ensemble model performs for all classes at a similar level in terms of prediction.

\begin{figure}[!t]
\centering
    \includegraphics[width=0.75\linewidth]{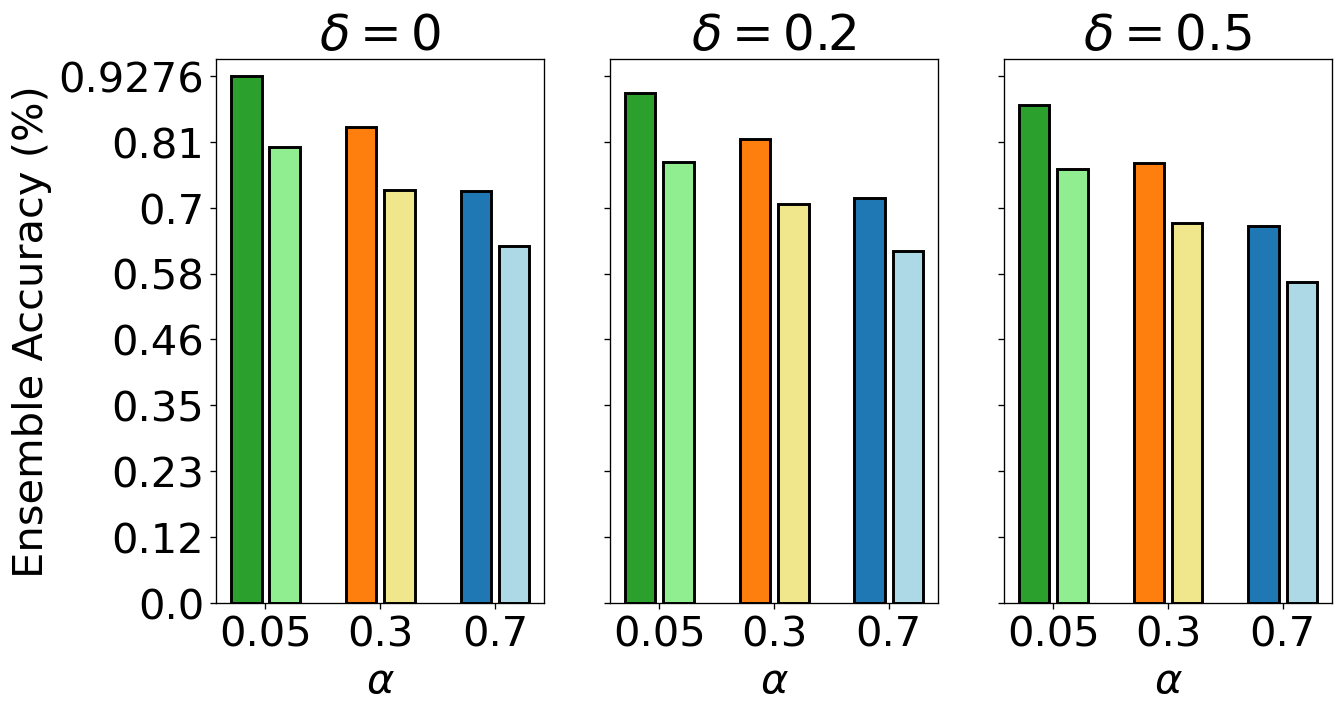}
    
  \caption[Ensemble Accuracy for Different Scenarios and Class Percentages]{Ensemble accuracy for different training scenarios and $\alpha$, $\delta$ values for CIFAR-10 dataset. Green, orange, and blue colors represent different $\alpha$ values, $0.05$, $0.3$, and $0.7$ respectively. Dark colors represent Scenario 1 and 3, and the light colors represent Scenario 2.}
  \label{image:ensembleAccuracies}
\end{figure}

In Figure~\ref{image:ensembleAccuracies}, the accuracy of the ensemble model for different choices of $\alpha$ and $\delta$ values and training scenarios are given for CIFAR-10. The figure shows that the ensemble accuracy increases as the $\alpha$ and $\delta$ values decrease. As expected, training with Scenario 1 and Scenario 3 yield better accuracy than Scenario 2. That means the model learns better as the heterogeneity of training data observed by edge devices increases, and the prediction performance is preserved as the heterogeneity in test data stays similar. It is also shown that the accuracy of the ensemble model is limited to the accuracy of the edge model. For CIFAR-10, the prediction accuracy of 91.14\%, 93.47\%, and 94.23\% is obtained using 20 edge models with a mean accuracy of 31.60\%, 37.46\%, and 48.41\%, respectively.

The effect of changing the number of edge models and the transferred feature vector size $L_{com}$ is also investigated in Table~\ref{table:ensembleModelChangingEdgeLcom} for CIFAR-10 and CIFAR-100 datasets. For changing the number of edge models, $L_{com}$ is taken as 64. For changing $L_{com}$, the number of edge models is taken as 20. It can be seen that increasing the number of edge models also increases the final ensemble accuracy. The same effect for changing $L_{com}$ can be observed, but the increase ceases to be material after a certain point of $L_{com}$ value, which is after 128 for this example.

\begin{table}[!t]
\centering
\caption{Test accuracy with different $N$ and $L_{com}$.}
\label{table:ensembleModelChangingEdgeLcom}
\resizebox{\columnwidth}{!}{
\begin{tabular}{|c|cccc|cccc|}
\hline
\multirow{2}{*}{\textbf{Dataset}} & \multicolumn{4}{c|}{\textbf{\boldsymbol{$N$}}}                                                                               & \multicolumn{4}{c|}{\textbf{$\boldsymbol{L_{com}}$}}                                                                                \\ \cline{2-9} 
                                  & \multicolumn{1}{c|}{\textbf{10}} & \multicolumn{1}{c|}{\textbf{20}} & \multicolumn{1}{c|}{\textbf{30}} & \textbf{50}    & \multicolumn{1}{c|}{\textbf{32}} & \multicolumn{1}{c|}{\textbf{64}} & \multicolumn{1}{c|}{\textbf{128}} & \textbf{256} \\ \hline
CIFAR-10                          & \multicolumn{1}{c|}{80.80}       & \multicolumn{1}{c|}{93.47}       & \multicolumn{1}{c|}{96.24}       & \textbf{98.79} & \multicolumn{1}{c|}{90.89}       & \multicolumn{1}{c|}{93.47}       & \multicolumn{1}{c|}{93.33}        & 93.47        \\ \hline
CIFAR-100                         & \multicolumn{1}{c|}{41.60}       & \multicolumn{1}{c|}{55.48}       & \multicolumn{1}{c|}{64.10}       & \textbf{75.77} & \multicolumn{1}{c|}{51.77}       & \multicolumn{1}{c|}{55.48}       & \multicolumn{1}{c|}{67.37}        & 67.40        \\ \hline
\end{tabular}
}
\end{table}

Comparison with recent edge intelligence studies regarding final accuracy for image classification and regression problems are presented in Table~\ref{table:accuracyComparison} and Table~\ref{table:regressionComparison}. The best-reported values are taken from the compared studies and our study. In the image classification comparison, accuracy results for CIFAR-10 and CIFAR-100 of EdgConvEns are taken from Table~\ref{table:ensembleModelChangingEdgeLcom}. MNIST and Fashion MNIST results of EdgConvEns are given for $N=20$ and $L_{com}=64$. In regression comparison, $N=50$ is taken for our results. Table~\ref{table:regressionComparison} shows that EdgeConvEns produces state-of-the-art accuracy for image classification and regression problems though it uses less accurate edge models than the ones used in the compared studies. For example, the edge models used in \cite{fedGKT} have 37.67\% and 78.94\% accuracy on CIFAR-100 and CIFAR-10, respectively. Table~\ref{table:edgeModelSpecifications} shows that the average edge model accuracy of EdgeConvEns is 8.43\% and 37.46\%, respectively.

\begin{table}[!t]
\centering
\caption{Accuracy comparison for image classification.}
\label{table:accuracyComparison}
\resizebox{\columnwidth}{!}{
\begin{tabular}{|c|c|c|c|c|}

\hline
\textbf{Method}                                                    & \textbf{CIFAR-10} & \textbf{CIFAR-100} & \textbf{MNIST}   & \textbf{\begin{tabular}[c]{@{}c@{}}Fashion\\ MNIST\end{tabular}} \\ \hline
FedAvg \cite{fedAvg}                                                             & 81.72\%           & -                  & 95.04\%          & 94.50\%                                                          \\ \hline
FedProx \cite{fedProx}                                                            & 83.25\%           & -                  & 96.26\%          & 94.53\%                                                          \\ \hline
FedDyn \cite{fedDyn}                                                             & 85.19\%           & 53.27\%            & -                & -                                                                \\ \hline
SCAFFOLD \cite{scaffold}                                                           & 85.99\%           & 53.32\%            & -                & -                                                                \\ \hline
FedGen \cite{fedGen}                                                             & 83.91\%           & 50.38\%            & -                & -                                                                \\ \hline
FedProto \cite{fedProto}                                                           & 84.49\%           & -                  & 97.13\%          & 97.10\%                                                          \\ \hline
FedFTG \cite{fedFTG}                                                             & 87.34\%           & 56.94\%            & -                & -                                                                \\ \hline
FedDC \cite{fedDC}                                                               & 86.18\%           & 55.52\%            & 98.45\%          & -                                                                \\ \hline
\begin{tabular}[c]{@{}c@{}}Bayesian\\ Nonparametric\end{tabular} \cite{fedBayesianNonparametric}   & 45.80\%           & -                  & 97.80\%          & -                                                                \\ \hline
AE-KD \cite{aekd}                                                              & 93.01\%           & 72.36\%            & -                & -                                                                \\ \hline
EKD \cite{ekd}                                                              & 92.33\%           & 67.78\%            & -                & -                                                                \\ \hline
FedGKT \cite{fedGKT}                                                            & 92.97\%           & 69.57\%            & -                & -                                                                \\ \hline
 \begin{tabular}[c]{@{}c@{}}Majority\\ Voting\end{tabular}                                                            & 42.30\%           & 14.06\%            &  93.12\%               & 82.74\%                                                                \\ \hline
\begin{tabular}[c]{@{}c@{}}Average\\ Voting\end{tabular}                                                             & 43.26\%           & 17.62\%            &    93.46\%             &  82.98\%                                                               \\ \hline

\textbf{EdgeConvEns}                                                            & \textbf{98.79\%}           & \textbf{75.77\%}            & \textbf{99.32\%}                & \textbf{97.40\%}                                                                \\ \hline
\end{tabular}
}
\end{table}

\begin{table}[!t]
\centering
\caption{Accuracy comparison for regression.}
\label{table:regressionComparison}

\resizebox{\columnwidth}{!}{
\begin{tabular}{|c|c|c|c|c|}
\hline
\textbf{Work}           & \textbf{Dataset}                                             & \textbf{Metric}      & \textbf{\begin{tabular}[c]{@{}c@{}}Reported\\ Value\end{tabular}} & \textbf{EdgeConvEns} \\ \hline
SAFA \cite{safa}        & \begin{tabular}[c]{@{}c@{}}Boston\\ Housing\end{tabular}     & \begin{tabular}[c]{@{}c@{}}Accuracy \cite{safa}\\(higher is better)\end{tabular} & 0.643                                                             & \textbf{0.728}                                                              \\ \hline
FedAG \cite{fedag}      & \begin{tabular}[c]{@{}c@{}}Boston\\ Housing\end{tabular}     & \begin{tabular}[c]{@{}c@{}}RMSE\\(lower is better) \end{tabular}                & 4.070                                                             & \textbf{3.479}                                                             \\ \hline
PrivFL \cite{privfl}    & \begin{tabular}[c]{@{}c@{}}Boston\\ Housing\end{tabular}     & \begin{tabular}[c]{@{}c@{}}MAE\\(lower is better)\end{tabular}                  & 3.266                                                             & \textbf{2.788}                                                              \\ \hline
Tiresias \cite{tiresias}      & \begin{tabular}[c]{@{}c@{}}California\\ Housing\end{tabular} & \begin{tabular}[c]{@{}c@{}}R2 Score\\(higher is better)\end{tabular}                   & 0.717                                                             & \textbf{0.726}                                                              \\ \hline
Sherpa \cite{sherpa}    & \begin{tabular}[c]{@{}c@{}}California\\ Housing\end{tabular} & \begin{tabular}[c]{@{}c@{}}R2 Score\\(higher is better)\end{tabular}              & 0.502                                                             & \textbf{0.868}                                                              \\ \hline
DER Forecast \cite{der} & Pecan Street                                                 & \begin{tabular}[c]{@{}c@{}}RMSE\\(lower is better) \end{tabular}                 & 1.98                                                              & \textbf{0.63}                                                               \\ \hline
\end{tabular}
}
\end{table}

\subsection{Memory and Communication Efficiency}
\label{subsection:memoryAndCommunicationEfficiency}

The average time required for the edge model training is 345, 402, and 500 minutes for $\alpha$ values of 0.05, 0.3, and 0.7, respectively. It can be seen that the edge models are trained in a shorter time when they are trained with less amount of data. In the whole training process, the training of the edge models takes the most time due to the hardware constraints of the edge devices.

\begin{figure}[!t]
\centering
  \subfigure[]{
    \includegraphics[width=.45\linewidth]{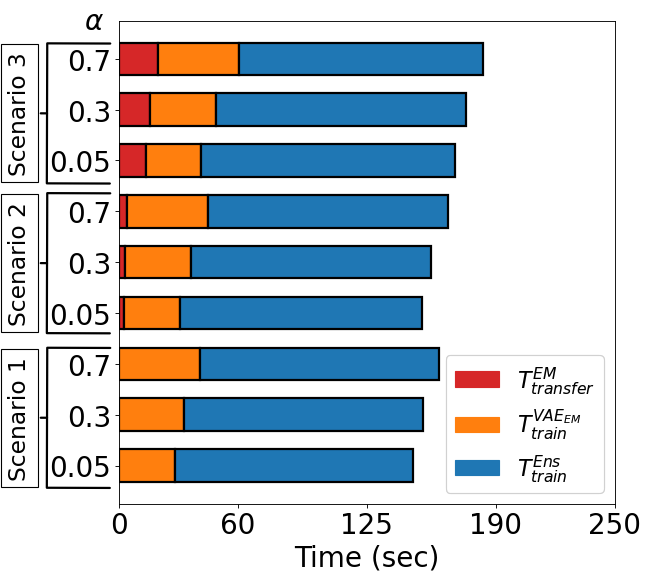}
    \label{image:timeRequirementsAndCumTransfer_Time}
    }
  \hspace{0.1pt}
  \subfigure[]{
    \includegraphics[width=.45\linewidth]{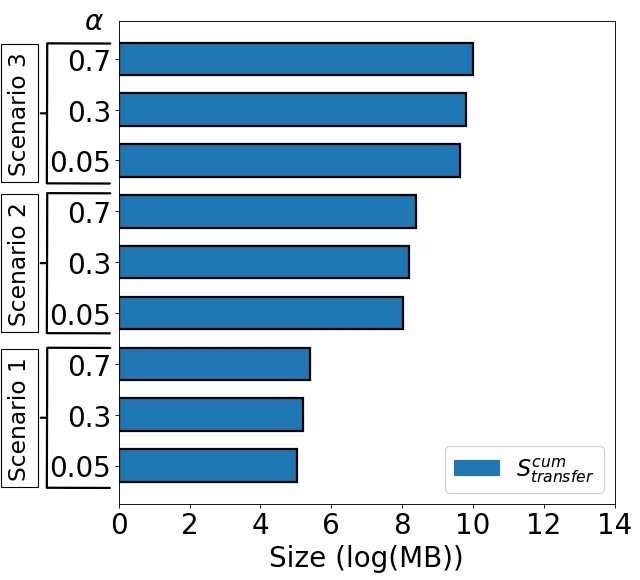}
    \label{image:timeRequirementsAndCumTransfer_CumTrns}
   }
  \caption{(a) Overall latency for feature vector transfer, VAE training, and
ensemble training, (b) Total cumulative transferred data amount from the edge.}
  \label{image:timeRequirementsAndCumTransfer}
\end{figure}

In Figure~\ref{image:timeRequirementsAndCumTransfer_Time}, the time requirements for feature vector transfer ($T^{EM}_{transfer}$), VAE training ($T^{VAE_{EM}}_{train}$), and ensemble model training ($T^{Ens}_{train}$) is shown. The results are invariant with the dataset used for training due to the same $L_{com}$ in all edge models. The total transfer time for an edge model is the most for Scenario 3 as a mini-batch is repeatedly transferred to the server. The number of communication is calculated as 39063 for Scenario 3. It is less in Scenario 2 due to the repeated use of the same mini-batch to complete the ensemble training epochs. The number of communication is calculated as 7813 for Scenario 2. It is the least for Scenario 1 because the feature vectors are sent to the server once. The ensemble learning time slightly increases from Scenario 1 to Scenario 3 due to repeated VAE decoding operations as the same mini-batch is received by the server multiple times. It can also be seen that the latency for feature vector transfer and VAE training visibly decreases for decreasing $\alpha$ values as expected. The cumulative amount of data ($S^{cum}_{transfer}$) of feature vectors transferred to the server is shown in Figure~\ref{image:timeRequirementsAndCumTransfer_CumTrns}. The largest amount is in Scenario 3 due to repeated transfers, while the least amount of feature vectors data transfer is in Scenario 1 due to one-time transfer. End-to-end latency is approximately 0.05 seconds for inference in all training scenarios.

Table~\ref{table:memoryRequirementsOnEdgeAndServer} shows the size of the transferred feature vectors in a one-way communication from one edge device to the server ($M^{EM}_{transfer}$), required storage memory on the server ($M^{server}_{memory}$), and required storage memory on an edge device ($M^{EM}_{memory}$) by training scenarios and $\alpha$ values. It shows that the largest memory is needed for Scenario 1 because the whole feature vector data of each edge model are transferred to the server once and stored there. However, the required memory remarkably reduces when the system is trained with Scenario 2 and Scenario 3. Scenario 3 requires no memory space allocated for the transferred feature vector data, whereas Scenario 2 requires it for only a mini-batch.

\begin{table}[!t]
\centering
\caption{Memory requirements on edge devices and server.}
\label{table:memoryRequirementsOnEdgeAndServer}
\begin{tabular}{|c|c|c|c|c|}
\hline
\textbf{Scenario} &
  \textbf{$\boldsymbol{\alpha}$} &
  \textbf{\begin{tabular}[c]{@{}c@{}}$\boldsymbol{M^{EM}_{transfer}}$\\ (MB)\end{tabular}} &
  \textbf{\begin{tabular}[c]{@{}c@{}}$\boldsymbol{M^{server}_{memory}}$\\ (MB)\end{tabular}} &
  \textbf{\begin{tabular}[c]{@{}c@{}}$\boldsymbol{M^{EM}_{memory}}$\\ (MB)\end{tabular}} \\ \hline
\multirow{3}{*}{Scenario 1} & 0.05                 & 7.7                 & \multirow{3}{*}{256.0} & 7.7                 \\ \cline{2-3} \cline{5-5} 
                            & 0.3                  & 9.0                 &                       & 9.0                 \\ \cline{2-3} \cline{5-5} 
                            & 0.7                  & 11.1                 &                       & 11.1                 \\ \hline
Scenario 2                  & \multirow{2}{*}{All} & \multirow{2}{*}{0.03} & 0.7                   & \multirow{2}{*}{0.03} \\ \cline{1-1} \cline{4-4}
Scenario 3                  &                      &                      & 0                     &                      \\ \hline
\end{tabular}
\end{table}

\section{Conclusion}
\label{section:conclusion}

This study proposes EdgeConvEns, a convolutional ensemble learning framework for deep edge intelligence. The EdgeConvEns framework aims to boost the system's overall classification and regression performance by learning a unified representation from the collective information extracted by weak edge models accessing heterogeneously distributed data. Thus, we target to reduce the computational requirements on edge while offering performances comparable to state-of-the-art. The proposed framework also tackles missing information due to failures in network communication by a VAE-based feature imputation approach. Moreover, EdgeConvEns provides a customizable acceleration for training DNNs on Xilinx FPGA devices. Thus, the training can be accelerated using different parallelization factors such that the hardware limits are met for different target devices. EdgeConvEns also makes the training of the system available for devices with differing hardware capacities, thanks to different learning scenarios. The experiments conducted with benchmark datasets and various training scenarios demonstrate that the proposed EdgeConvEns outperforms conventional ensemble learning and standard federated learning techniques.

\bibliographystyle{IEEEtran}
\bibliography{references}

\end{document}